\documentclass[sigconf]{acmart}
\usepackage{tikz}
\usetikzlibrary{positioning}

\usepackage{booktabs}  
\usepackage{siunitx}   
\usepackage{xcolor}    
\usepackage{tikz}
\usetikzlibrary{positioning, shadows, decorations.pathreplacing, backgrounds}
\usepackage{booktabs}
\usepackage{siunitx}
\usepackage{colortbl}
\usetikzlibrary{positioning, arrows.meta}

\usepackage{tikz}
\usetikzlibrary{patterns}
\usepackage{amsmath}
\usepackage[multiple]{footmisc}
\usepackage{booktabs}
\usepackage{amsthm}
\usepackage{dsfont}

\usepackage[utf8]{inputenc}
\usepackage{balance}
\usepackage{siunitx}
\usepackage{multirow}
\usepackage{graphicx}
\usepackage{listings}
\usepackage{commath}
\usepackage{epstopdf}
\usepackage{color}
\usepackage{url}
\usepackage{caption}
\usepackage{alltt}
\usepackage{bbm}
\usepackage{mathrsfs}
\usepackage{float}
\usepackage{subcaption}
\usepackage{graphicx,fancyvrb}
\usepackage[linesnumbered,ruled,vlined]{algorithm2e}
\usepackage{algpseudocode}

\usepackage{mathtools}
\usepackage{capt-of}
\usepackage{colortbl}
\usepackage{xcolor}
\usepackage{xfrac}
\usepackage{footnote}
 \usepackage{enumitem}
\usepackage{array}
\usepackage{arydshln}
\usepackage{cleveref}
\definecolor{mygreen}{rgb}{0,0.6,0}
\definecolor{myred}{rgb}{0.6,0,0}
\definecolor{mygray}{rgb}{0.5,0.5,0.5}
\definecolor{mymauve}{rgb}{0.58,0,0.82}
\definecolor{myblue}{rgb}{0,0,1}

\usepackage{listings}
\lstnewenvironment{VerbatimText}[1][]{
    
    \lstset{fancyvrb=true,basicstyle=\footnotesize,captionpos=b,xleftmargin=2em,#1}
}{}

\PassOptionsToPackage{hyphens}{url}\usepackage{hyperref}

\newcommand{\Dom}{\mathcal}

\DeclareMathAlphabet\mathbfcal{OMS}{cmsy}{b}{n}

\DeclareMathOperator*{\argmin}{argmin}

\definecolor{ReviewerOneColor}{RGB}{240, 110, 0}   
\definecolor{ReviewerTwoColor}{RGB}{0, 139, 109}   
\definecolor{ReviewerThreeColor}{RGB}{45, 85, 205}
\definecolor{MetaReviewerColor}{RGB}{142, 69, 133} 

\definecolor{darkred}{rgb}{0.55, 0.0, 0.0}

\newcommand{\mhm}[1]{\textcolor{darkgreen}{Mohammad-Hossein: #1}}
\newcommand{\nit}[1]{\textit{#1}}

\newcommand{\starter}[1]{\vspace{2mm}\noindent {\bf #1}}

\newcommand{\AUC}{\text{AUC}\xspace}
\newcommand{\xAUC}{\text{xAUC}\xspace}
\newcommand{\pr}{\text{Pr}\xspace}
\newcommand{\TPR}{\text{TPR}\xspace}
\newcommand{\FPR}{\text{FPR}\xspace}
\newcommand{\PR}{\text{PR}\xspace}
\newcommand{\EOD}{\text{EOD}\xspace}
\newcommand{\EO}{\text{EO}\xspace}
\newcommand{\DP}{\text{DP}\xspace}
\newcommand{\DSP}{\text{DSP}\xspace}

\newcommand{\ditto}{\texttt{Ditto}\xspace}
\newcommand{\hiergat}{\texttt{HierGAT}\xspace}
\newcommand{\dmatcher}{\texttt{DeepMatcher}\xspace}
\newcommand{\hmatcher}{\texttt{HierMatcher}\xspace}
\newcommand{\magellan}{\texttt{Magellan}\xspace}

\newcommand{\walamz}{\texttt{WAL-AMZ}\xspace}
\newcommand{\amzgog}{\texttt{AMZ-GOG}\xspace}
\newcommand{\DBLPGogS}{\texttt{DBLP-GOGS}\xspace}
\newcommand{\DBLPACM}{\texttt{DBLP-ACM}\xspace}

\newcommand{\dom}[1]{\Dom #1}

\newcommand{\ignore}[1]{}

\definecolor{black}{rgb}{0,0,0}
\definecolor{grey}{rgb}{0.8,0.8,0.8}
\definecolor{red}{rgb}{1,0,0}
\definecolor{green}{rgb}{0,1,0}
\definecolor{darkgreen}{rgb}{0,0.5,0}
\definecolor{darkpurple}{rgb}{0.5,0,0.5}
\definecolor{darkdarkpurple}{rgb}{0.3,0,0.3}
\definecolor{blue}{rgb}{0,0,1}
\definecolor{shadegreen}{rgb}{0.95,1,0.95}
\definecolor{shadeblue}{rgb}{0.95,0.95,1}
\definecolor{shadered}{rgb}{1,0.85,0.85}
\definecolor{shadegrey}{rgb}{0.85,0.85,0.85}
\definecolor{oddRowGrey}{rgb}{0.80,0.80,0.80}
\definecolor{evenRowGrey}{rgb}{0.85,0.85,0.85}
\definecolor{lightpurple}{rgb}{0.88,1.0,1.0}

\newcommand{\RNum}[1]{\uppercase\expandafter{\romannumeral #1\relax}}

\newtheorem{defn}{Definition}[section]

\newtheorem{col}[defn]{Colollary}

\newcommand{\proj}[1]{{\Pi}}
\newcommand{\sel}[1]{{\sigma}}

\newcommand{\cut}[1]{}
\newcommand{\eat}[1]{}

\usepackage[multiple]{footmisc}

\SetKwInput{KwInput}{Input}
\SetKwInput{KwOutput}{Output}

\usepackage{siunitx}
\ExplSyntaxOn
  \cs_new_eq:NN \calc \fp_eval:n
\ExplSyntaxOff

\definecolor{cbgreen}{RGB}{0,158,115} 
\definecolor{cbred}{RGB}{213,94,0} 

\newcommand{\stockupGreen}[1]{%
    \tikz{\filldraw[fill=cbgreen, draw=none] (0,0) -- (0.2cm,0) -- (0.1cm,0.2cm) -- cycle;} #1%
}

\newcommand{\stockupRed}[1]{%
    \tikz{\filldraw[fill=cbred, draw=none] (0,0) -- (0.2cm,0) -- (0.1cm,0.2cm) -- cycle;} #1%
}

\newcommand{\stockdown}[1]{%
    \tikz{\filldraw[fill=cbgreen, draw=none] (0,0) -- (0.2cm,0) -- (0.1cm,-0.2cm) -- cycle;} #1%
}

\newcommand{\stockdownRed}[1]{%
    \tikz{\filldraw[fill=cbred, draw=none] (0,0) -- (0.2cm,0) -- (0.1cm,-0.2cm) -- cycle;} #1%
}

\definecolor{DodgerBlue}{RGB}{30, 144, 255}
\newcommand{\stocknochange}[1]{%
    \tikz{\filldraw[fill=DodgerBlue, draw=none] (0,0) rectangle (0.15cm,0.15cm);} #1%
}

\AtBeginDocument{%
  \providecommand\BibTeX{{%
    \normalfont B\kern-0.5em{\scshape i\kern-0.25em b}\kern-0.8em\TeX}}}

\usepackage{setspace}

\renewcommand\footnotetextcopyrightpermission[1]{} 

\title{Threshold-Independent Fair Matching through Score Calibration}

\author{Mohammad Hossein Moslemi}
\affiliation{%
  \institution{University of Western Ontario}
  \city{London}
  \country{Canada}
}
\email{mohammad.moslemi@uwo.ca}
\orcid{0009-0002-0278-4665}
\authornotemark[1]

\author{Mostafa Milani}
\affiliation{%
  \institution{University of Western Ontario}
  \city{London}
  \country{Canada}
}
\email{mostafa.milani@uwo.ca}
\orcid{0000-0002-3386-7079}

\begin{document}

\pagenumbering{Roman} 

\pagenumbering{arabic}

\begin{abstract}
Entity Matching (EM) is a critical task in numerous fields, such as healthcare, finance, and public administration, as it identifies records that refer to the same entity within or across different databases. EM faces considerable challenges, particularly with false positives and negatives. These are typically addressed by generating matching scores and apply thresholds to balance false positives and negatives in various contexts.However, adjusting these thresholds can affect the fairness of the outcomes, a critical factor that remains largely overlooked in current fair EM research. The existing body of research on fair EM tends to concentrate on static thresholds, neglecting their critical impact on fairness. To address this, we introduce a new approach in EM using recent metrics for evaluating biases in score-based binary classification, particularly through the lens of {\em distributional parity}. This approach enables the application of various bias metrics—like equalized odds, equal opportunity, and demographic parity—without depending on threshold settings. Our experiments with leading matching methods reveal potential biases, and by applying a calibration technique for EM scores using {\em Wasserstein barycenters}, we not only mitigate these biases but also preserve accuracy across real-world datasets. This paper contributes to the field of fairness in data cleaning, especially within EM, which is a central task in data cleaning, by promoting a method for generating matching scores that reduce biases across different thresholds.


\end{abstract}
\maketitle

\section{Introduction}\label{sec:intro}

Entity matching (EM), also known as entity linkage or record matching, plays a crucial role in various applications across multiple domains, including healthcare, finance, e-commerce, and public administration. It identifies records referring to the same real-world entity across different or single databases, integrating fragmented information, improving data quality, and providing a unified view of data entities. Despite extensive research and numerous matching methods, challenges in entity matching (EM) due to data heterogeneity continue to cause false positives and false negatives across domains and datasets. These errors can introduce biases and discrimination, as highlighted in recent research~\cite{shahbazi2023through,nilforoushan2022entity,efthymiou2021fairer,louis2022online,makri2022towards}. This short paper discusses overlooked biases in state-of-the-art EM methods related to fairness.

EM methods often generate a {\em matching score} and use a {\em matching threshold} to derive a binary decision. This process inherently allows for the tuning of thresholds across different applications to balance between false positives and false negatives. However, for varying matching thresholds, biases in matching must be evaluated independent of a particular threshold. Previous research on fairness in EM has often overlooked the significant effects of changing thresholds~\cite{efthymiou2021fairer,louis2022online,makri2022towards}, concentrated on specific thresholds~\cite{shahbazi2023through}, or used metrics that can be misleading for evaluating bias in a threshold-independent manner~\cite{nilforoushan2022entity}.

\begin{figure}
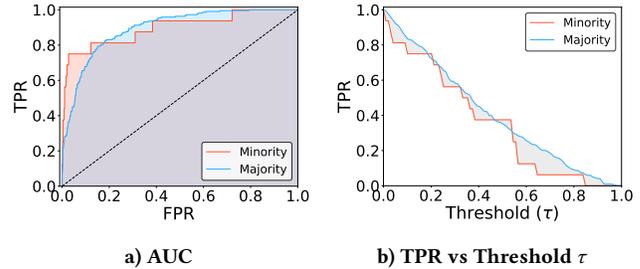

         \centering\subfloat[AUC]{\label{fig:auc-two}
			{\includegraphics[width=0.23\textwidth]{fig/Intro_AUC.pdf}}}
		\hfill\subfloat[TPR vs Threshold $\tau$]{\label{fig:tpr}
			{\includegraphics[width=0.23\textwidth]{fig/Intro_EQ_OPP.pdf}}}	
			\vspace{-3mm}
	\caption{The ROC of \magellan~\cite{konda2018magellan} (Figure~\ref{fig:auc-two}) shows similar AUC, 0.897 and 0.894 while there is a significant performance difference in certain thresholds. Figure~\ref{fig:tpr} demonstrates a significant difference in \TPR at different thresholds.}\label{fig:combined}
\end{figure}

\begin{example}\em 
Adjusting matching thresholds in EM applications is key to optimizing matching performance. For example, using lower thresholds increases the chances of identifying individuals on a no-fly list, accepting the risk of some wrongful identifications to ensure criminals are not missed. Conversely, in financial services, a higher threshold is used to avoid wrongly linking and merging accounts due to the significant privacy and financial risks involved; here, it's acceptable to miss some matches. Likewise, in healthcare, setting a higher threshold is crucial to prevent the merging of wrong patient records, which helps avoid treatment mistakes, incorrect diagnoses, and life-threatening treatment errors. However, changing thresholds can affect fairness. A method that seems fair at one threshold might show significant biases at another, as adjusting the threshold can create or increase biases. To illustrate, the ROC curves in Figure~\ref{fig:auc-two} display the true positive and false positive rates (\TPR and \FPR) across various thresholds for minority and majority groups in the matching method \magellan~\cite{konda2018magellan} using the Amazon-Google dataset~\cite{mudgal2018deep}. Additionally, Figure~\ref{fig:tpr} demonstrates the \TPR for these groups at different thresholds. These figures show that although the matching method can perform similarly for both groups at certain thresholds (where the curves for the two groups intersect), at other thresholds, significant biases emerge, as shown by the differences in the curves.\end{example}

Recent work in~\cite{nilforoushan2022entity} explores threshold-independent fairness in EM, following the AUC-based bias metrics for binary classifiers in~\cite{kallus2019fairness,yang2023minimax,vogel2021learning}. These metrics use the AUC gap between minority and majority groups to measure bias. However, these bias measures have been criticized for potential ambiguity or misleading interpretations, as pointed out by Kwegyir et al.~\cite{kwegyir2023misuse}. The critique is based on the fact that AUC, as an aggregate measure, may not capture specific threshold biases. For instance, Figure~\ref{fig:auc-intro} illustrates that although the AUC values for both groups are almost identical, the ROC curves significantly differ at certain thresholds, revealing biases at those thresholds.

\ignore{\begin{figure}
    \centering
\begin{subfigure}[b]{0.44\columnwidth}
    \centering
    \includegraphics[width=\textwidth]{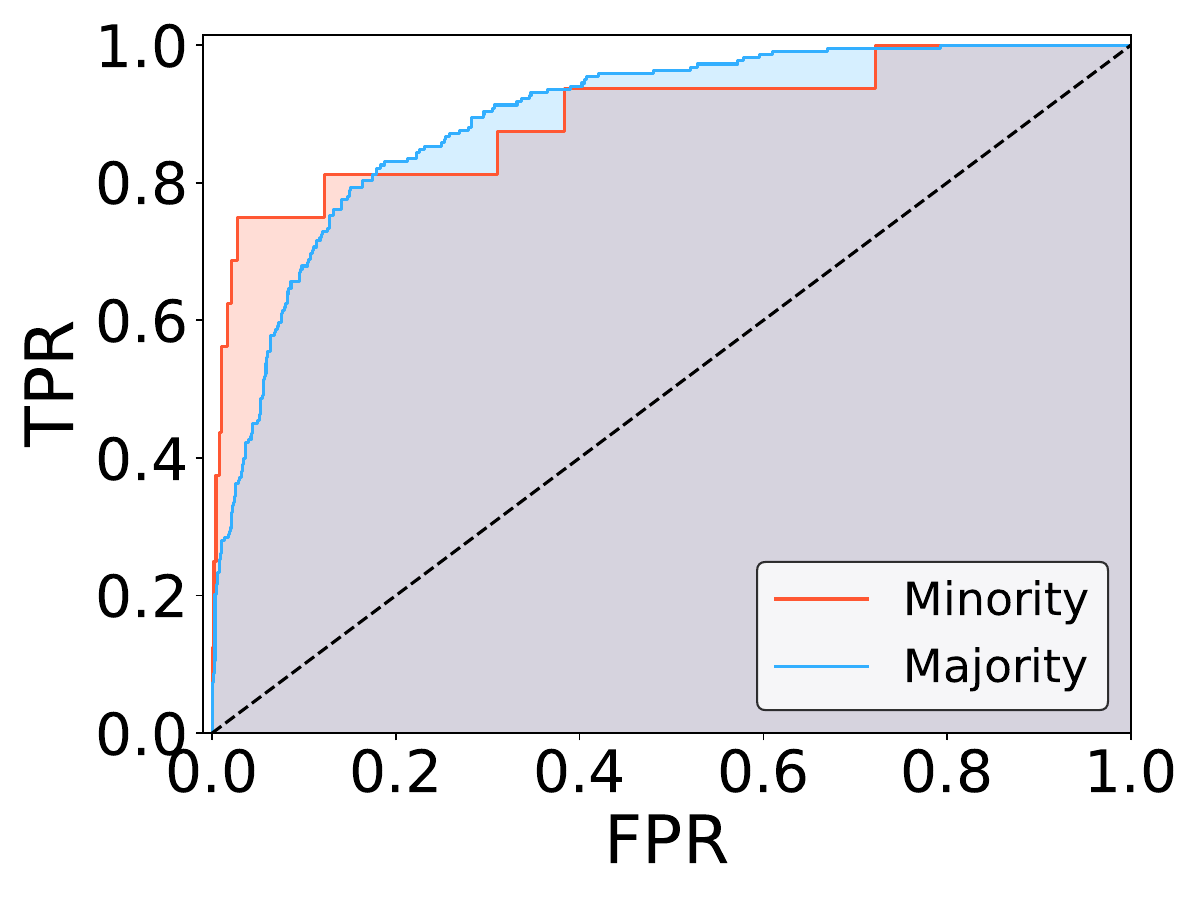}
        \caption{AUC}
        \label{fig:auc}
    \end{subfigure}
    \quad
\begin{subfigure}[b]{0.44\columnwidth}
        \centering  
        \includegraphics[width=\textwidth]{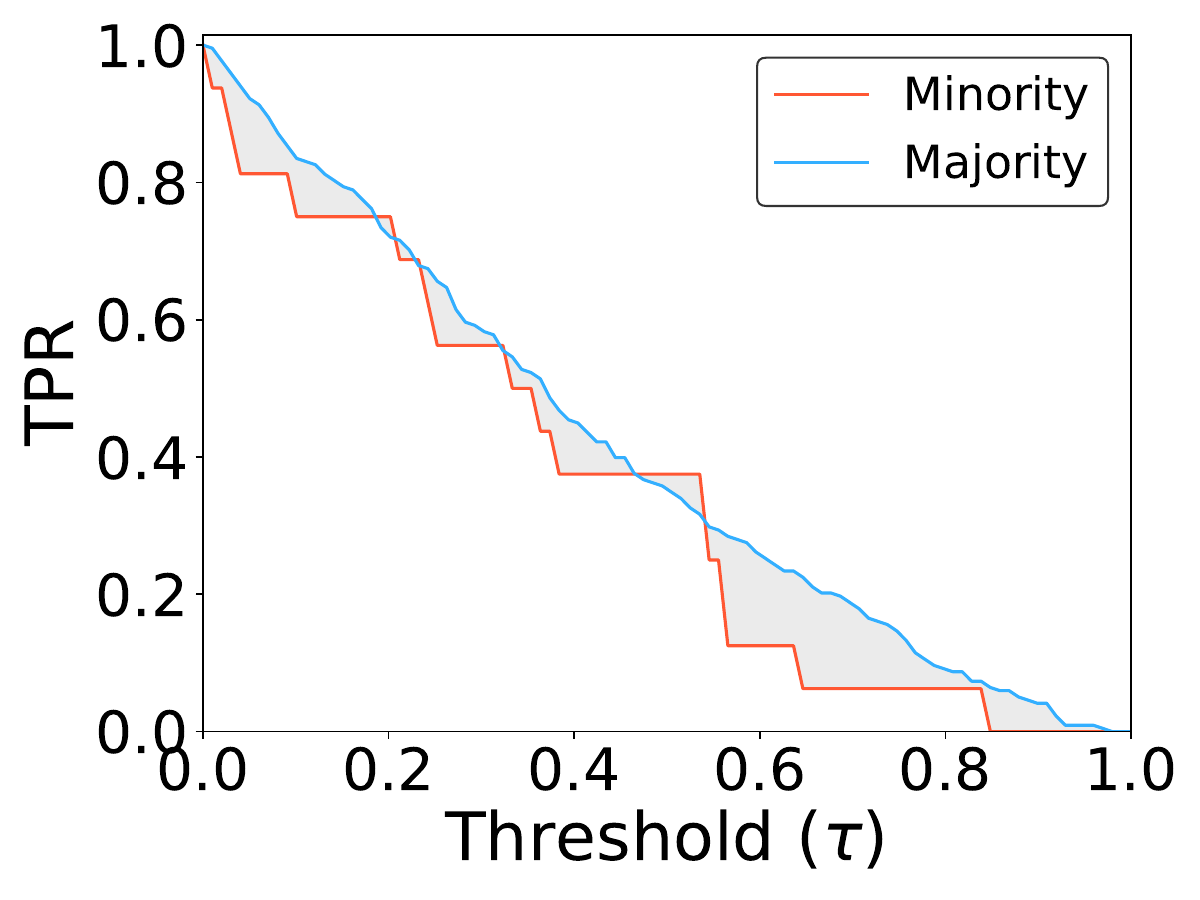}
        \caption{TPR vs Threshold $\tau$}
        \label{fig:tpr}
    \end{subfigure}
    \caption{Similar AUC values of 0.897 and 0.894 are shown in Figure~\ref{fig:auc}. However, Figure~\ref{fig:tpr} demonstrates a significant difference in TPR at different thresholds.}
    \label{fig:combined}
\end{figure}}

To overcome the shortcomings highlighted in previous EM fairness research, we employ the recent metrics for evaluating biases in score-based binary classification without depending on a specific threshold~\cite{kwegyir-aggrey2023repairing,jiang2020wasserstein,gouic2020projection,chzhen2020fair}. We use {\em distributional parity} (\DSP), which limits the expected value of the difference in measures such as positive rate (\PR), \TPR, or \FPR between groups. The gray area in Figure~\ref{fig:tpr} illustrates this for TPR. Using \PR, \TPR, or both \TPR and \FPR, \DSP extends the principle of demographic parity (\DP), equal opportunity (\EO), and equalized odds (\EOD) across various thresholds.


Through \DSP, we examine state-of-the-art matching techniques to uncover biases, then formulate fair EM by calibrating and adjusting matching score distributions across groups. Using {\em Wasserstein barycenters} from optimal transport (OT) theory, this post-processing fairness approach corrects biases without retraining the initial method. Our experiments show that this calibration reduces biases measured by \DSP while maintaining accuracy.

We provide an overview of the relevant background, formulate the matching problem, and explain distributional parity within the context of EM in \S\ref{sec:background}. The calibration method is detailed in \S\ref{sec:methods}, \S\ref{sec:experiment} presents our experimental results, and \S\ref{sec:conclusion} concludes the paper.
\section{Background and Problem Definition} \label{sec:background}

Let $f: \Dom{X}\times \Dom{A} \mapsto \{0,1\}$ be a binary classifier where $\Dom{X}$ is a feature space, and $\Dom{A}=\{a,b\}$ is a set of binary protected (sensitive) attribute values with $a$ and $b$ representing the majority and minority groups, respectively. We assume $f$ is derived from a score function $s: \Dom{X}\times \Dom{A} \mapsto [0,1]$ and a classification threshold $\tau \in [0,1]$, where $f(r)=\mathbbm{1}_{s(r) \ge \tau}$ for every $r\in \Dom{X}\times \dom{A}$ ($\mathbbm{1}$ is the indicator function). We use $f_a$ and $f_b$ ($s_a$ and $s_b$) to refer to the classifier (the score function) within majority and minority groups.

By employing a probability distribution ``$\pr$'' with random variables ${\bf X}$, $A$, and $Y$, which denote the features, the protected attribute, and the true label, respectively, we define the positive rate (\PR) within a specific group $g\in \Dom{A}$ as $\PR_g(s, \tau)=\pr(s(X)\ge \tau \mid A=g)$. Similarly, we define \TPR and \FPR for each group. From \PR, we derive \DP as $|\PR_a(s, \tau)-\PR_b(s, \tau)|\le \epsilon$, where $\epsilon$ represents a limit on disparity or bias. \TPR and \FPR are used to measure fairness in terms of \EO and \EOD; for \EO, we have $|\TPR_a(s, \tau)-\TPR_b(s, \tau)|\le \epsilon$, and for \EOD, both $|\TPR_a(s, \tau)-\TPR_b(s, \tau)|\le \epsilon$ and $|\FPR_a(s, \tau)-\FPR_b(s, \tau)|\le \epsilon$ are required.

The AUC for the scoring function $s$ is expressed as $\nit{AUC}(s)=\int_{0}^{1} \TPR(s, \FPR^{-1}(s, \tau))\;d \tau$. It quantifies the likelihood that a positive record is accurately ranked above a negative one, offering a measure of accuracy that does not depend on any specific threshold $\tau$. AUC-based fairness introduces a threshold-independent fairness criterion inspired by the concept of cross-group AUC (xAUC) between two groups $a$ and $b$, defined as $\xAUC_a^b(s)=\int_{0}^{1} \TPR_a(s, \FPR_b^{-1}(s, \tau))$ $d \tau$~\cite{kallus2019fairness}. This effectively measures the chance that a positively labeled record from group $a$ is ranked above a negatively labeled record from group $b$. Fairness in AUC terms is captured by $\Delta \xAUC(s)=\xAUC_a^b(s)-\xAUC_b^a(s) \le \epsilon$, reflecting the bias or disparity in scoring between the groups, $a$ and $b$.

\starter{Distributional Parity~\cite{kwegyir-aggrey2023repairing}:} Distributional parity, or \DSP, extends \DP, \EO, and \EOD to define threshold-independent fairness. Intuitively, it measures these traditional measures across varying thresholds by computing their expected value. More precisely, for any measure $\gamma \in \{\PR, \TPR, \FPR\}$, \DSP is achieved when $\mathbb{E}_{\tau\sim [0,1]}|\gamma_a(\tau)-\gamma_b(\tau)|=0,$ where $\mathbb{E}_{\tau\sim [0,1]}$ is the expected value for iid sampled thresholds $\tau$ from $[0,1]$. When the expected value is nonzero, we denote it by \(\mathcal{U}(s, \gamma)\) as a measure of bias and refer to it as \DSP.

\starter{The Fair EM Problem:} In EM, a matcher acts as a binary classifier within a feature space $\mathcal{X}=\mathcal{E}\times \mathcal{E}$, where $\mathcal{E}$ is the feature space for individual records. Each record would have its own protected attribute. A record pair is defined as a minority pair if both records belong to the minority group and a majority pair if both belong to the majority group. Combinations of one minority and one majority record are excluded, as these are identified as non-matches by blocking and removed before classification. The matching score, denoted as $s:\mathcal{E}^2\times \mathcal{A} \rightarrow [0,1]$, and other performance metrics like \PR, \TPR, \FPR, and DSP are defined as in prior discussions, using the protected attribute $\mathcal{A}$, which is the same in both records.

Previous research~\cite{nilforoushan2022entity,shahbazi2023through} has shown that current matching techniques are biased. We plan to assess these techniques using \DSP to identify biases without specific thresholds. Our findings reveal distributional disparity, indicating uneven performance across groups. To address these biases, we introduce fair EM, which adjusts scores from current methods to meet a broad fairness criterion.

\begin{definition} \label{df:fair} \em Given a score function $s$ and a fairness constraint $\sigma:\ignore{\mathbb{E}_{\tau\sim [0,1]}|\gamma_a(\tau)-\gamma_b(\tau)|=}\mathcal{U}(s, \gamma) \le \epsilon$, where $\gamma$ is any measure in $\{\PR, \TPR, \FPR\}$, the problem of fair EM is to find a score $s^*$ that satisfies $\sigma$ and minimally differs from $s$, i.e., it minimizes  $\delta(s^*,s)=\mathbb{E}_{r\sim \Dom{E}^2 \times \Dom{A}}|s^*(r)-s(r)|$.\end{definition}

The objective of fair EM is to find a fair score function, \(s^*\), that closely resembles the original matching score, \(s\), and therefore enjoys high accuracy while adhering to the fairness constraint, \(\sigma\), in terms of \DSP with respect to \(\gamma\).

\section{Score Calibration using Barycenter} \label{sec:methods}

\begin{table*}[h!]\centering
    \resizebox{0.9\textwidth}{!}{
    \begin{tabular}{llllllllllll}\\
    \toprule
     \multirow{2}{*}{\rotatebox[origin=c]{90}{\small Dataset}} & \multirow{2}{*}{Method}  & \multicolumn{3}{c}{DSP (before)} & \multicolumn{2}{c}{before} & \multicolumn{3}{c}{DSP (after)} & \multicolumn{2}{c}{after} \\
    \cmidrule(lr){3-5} \cmidrule(lr){6-7} \cmidrule(lr){8-10} \cmidrule(lr){11-12} 
     & & \DP & \EO & \EOD & AUC & $\Delta\xAUC$ & \DP & \EO & \EOD & AUC & $\Delta\xAUC$ \\
    \midrule
\multirow{4}{*}{\rotatebox[origin=c]{90}{ \small \amzgog~}}
& \ditto & 7.17 & 12.40 & 14.71 & 95.95 & 2.26 & 2.37 \stockdown{4.8} & 5.83 \stockdown{6.57} & 6.46 \stockdown{8.24} & 95.86 \stockdownRed{0.09} &  2.23 \stockdown{0.03}\\
& \hiergat & 11.10 & 17.33 & 21.57 & 96.75 & 0.41 & 6.64 \stockdown{4.45} & 11.60 \stockdown{5.73} & 12.20 \stockdown{9.37} & 94.91 \stockdownRed{1.84} &  9.13 \stockupRed{8.72}\\
& \dmatcher & 8.53 & 11.74 & 15.22 & 84.99 & 5.62 & 3.46 \stockdown{5.06} & 10.34 \stockdown{1.39} & 11.19 \stockdown{4.03} & 84.64 \stockdownRed{0.35} &  5.68 \stockupRed{0.06}\\
& \hmatcher & 8.64 & 17.58 & 20.83 & 94.23 & 0.61 & 5.03 \stockdown{3.61} & 8.03 \stockdown{9.56} & 8.35 \stockdown{12.48} & 93.93 \stockdownRed{0.3} &  0.87 \stockupRed{0.26}\\
\midrule

\multirow{4}{*}{\rotatebox[origin=c]{90}{ \small \walamz~}}
& \ditto & 3.35 & 12.69 & 12.29 & 96.16 & 1.25 & 2.51 \stockdown{0.84} & 9.72 \stockdown{2.97} & 9.76  \stockdown{2.53} & 93.79 \stockdownRed{2.36} &  3.91 \stockupRed{2.66}\\
& \hiergat & 2.59 & 14.64 & 13.45 & 96.00 & 3.06 & 2.59 \stocknochange & 14.64  \stocknochange & 13.45  \stocknochange & 96.0 \stocknochange &  3.06 \stocknochange\\
& \dmatcher & 3.57 & 31.80 & 31.22 & 83.35 & 5.86 & 3.44 \stockdown{0.13} & 16.10 \stockdown{15.7} & 16.73 \stockdown{14.49} & 83.34 \stockdownRed{0.01} &  5.9 \stockupRed{0.04}\\
& \hmatcher & 2.15 & 11.99 & 12.08 & 94.06 & 5.21 & 2.15 \stocknochange & 11.99 \stocknochange & 12.08 \stocknochange & 94.06 \stocknochange &  5.21 \stocknochange\\
\midrule

\multirow{4}{*}{\rotatebox[origin=c]{90}{ \small \DBLPGogS~}}
& \ditto & 4.17 & 2.19 & 2.62 & 99.43 & 0.44 & 4.00 \stockdown{0.18} & 1.99 \stockdown{0.2} & 2.62 \stocknochange & 99.43 \stocknochange &  0.44 \stocknochange\\
& \hiergat & 4.55 & 3.46 & 2.97 & 99.68 & 0.34 &  3.41 \stockdown{1.14} & 1.17 \stockdown{2.29} & 1.53 \stockdown{1.44} & 99.68 \stocknochange &  0.26 \stockdown{0.09}\\
& \dmatcher & 5.71 & 8.61 & 8.32 & 98.58 & 4.26 & 4.71 \stockdown{1.0} & 8.32 \stockdown{0.29} & 6.96 \stockdown{1.36} & 98.59 \stockupGreen{0.01} &  4.32 \stockupRed{0.05}\\
& \hmatcher & 4.76 & 3.64 & 3.66 & 99.59 & 0.64 & 3.36 \stockdown{1.4} & 3.28 \stockdown{0.36} & 3.51 \stockdown{0.15} & 99.59 \stocknochange &  0.65 \stockupRed{0.01}\\
\midrule

\multirow{4}{*}{\rotatebox[origin=c]{90}{ \small \DBLPACM~}}
& \ditto & 1.06 & 0.55 & 0.52 & 99.88 & 0.08 & 1.06 \stocknochange & 0.55 \stocknochange & 0.52 \stocknochange & 99.88 \stocknochange &  0.08 \stocknochange\\
& \hiergat & 0.94 & 1.24 & 0.94 & 99.99 & 0.03 & 0.82 \stockdown{0.11} & 0.79 \stockdown{0.45} & 0.64 \stockdown{0.3} & 99.99 \stocknochange &  0.03 \stocknochange\\
& \dmatcher & 0.64 & 0.51 & 0.54 & 99.94 & 0.18 & 0.64 \stocknochange & 0.51 \stocknochange & 0.54 \stocknochange & 99.94 \stocknochange &  0.18 \stocknochange\\
& \hmatcher & 0.81 & 1.05 & 1.19 & 99.96 & 0.05 & 0.81 \stocknochange & 1.05 \stocknochange & 1.19 \stocknochange & 99.96 \stocknochange &  0.05 \stocknochange\\
    \bottomrule
    \end{tabular}
    }
    \caption{Bias (\DP, \EO, \EOD, $\Delta\xAUC$) and accuracy (\AUC) for matching methods (\ditto, \hiergat, \dmatcher, \hmatcher) over benchmarks (\amzgog, \walamz, \DBLPGogS, \DBLPACM), evaluated before and after score calibration. Values are scaled by $\times 100$. \protect\stocknochange means no change after calibration. The color green shows improvement, for instance, an increase in accuracy (\protect\stockupGreen\!) or a reduction in bias (\protect\stockdown\!). Orange denotes deterioration, such as a decrease in accuracy (\protect\stockdownRed\!) or a rise in bias (\protect\stockupRed\!).}
        \label{tab:bias}
\end{table*}

In this paper, we introduce a novel method for fair EM by calibrating scores from $s$ using the Wasserstein barycenter, derived from optimal transport theory. This method was previously suggested to improve fairness in~\cite{chzhen2020fair}. As an initial solution to fair EM in Definition~\ref{df:fair}, our method doesn't fully meet all fairness constraints or promise a minimal change from the original scores defined in the definition. However, our experiments show it can reduce bias without losing much score accuracy. We postpone developing a more detailed solution to our future work and discuss. First, we explain the basics of the Wasserstein distance and barycenter.


In OT theory, the Wasserstein distance\ignore{, also known as the Earth Mover's Distance (EMD),} between two probability measures $\mu_1$ and $\mu_2$ over a metric space is defined as:
\begin{equation}
W^p_p(\mu_1, \mu_2) = \inf_{\pi \in \Pi(\mu_1, \mu_2)} \int_{M \times M} d(x, y)^p \, d\pi(x, y),
\end{equation}
where $p \geq 1$ is the order of the distance, $d(x, y)$ is the cost of moving between points $x$ and $y$ in the metric space, and $\Pi(\mu_1, \mu_2)$ represents the set of all joint distributions $\pi$ in the metric space with marginals $\mu_1$ and $\mu_2$. The Wasserstein barycenter $\mu_1$ and $\mu_2$ with respect to weights $\alpha$ is defined as the measure $\hat{\mu}$ that minimizes the weighted sum of the Wasserstein distances to $\mu_1$ and $\mu_2$ in $\hat{\mu} = \argmin_{\mu} \left( \alpha W_p^p(\mu_1, \mu) + (1-\alpha) W_p^p(\mu_2, \mu) \right).$ This barycenter can be interpreted as a generalized mean of the measures $\mu_1$ and $\mu_2$ under the Wasserstein distance metric.

Our fair EM method treats the scoring functions \(s_a\) and \(s_b\) for groups $a$ and $b$ as distributions that are moved towards a common central Wasserstein barycenter, \(\hat{s}\). This alignment reduces unfairness by making score distributions more similar across groups. To ensure fairness in EM with respect to \DP, the score distributions for both minority and majority groups should be similar. This approach, based on barycenters, was chosen for its promising results and balance of fairness and accuracy. We plan to conduct an extensive search for alternative calibration methods in future research.


Moving scores to the barycenter, \(\hat{s}\), does not always result in the best scoring function. For instance, when $\gamma$ is \PR, \(\hat{s}\) makes \PR even for all groups at every level, which meets the fairness goals for \DP and \DSP (meaning \(\mathcal{U}(\hat{s},\text{PR})=0\)). Yet, this does not ensure the scores are as close as possible to the original scores $s$. When $\gamma$ is either \TPR or \FPR, using \(\hat{s}\) may not lead to the best scores. Unlike with $\gamma=$ \PR, \(\hat{s}\) does not reduce \(\mathcal{U}(\hat{s},\gamma)\) to its lowest possible value because \TPR and \FPR depend on the actual labels, not just on the distribution of scores as \PR does.

This means that just moving $s_a$ and $s_b$ directly to the barycenter $\hat{s}$ might not result in the best score, like $s^*$ mentioned in Definition~\ref{df:fair}. To fix this issue, ~\cite{kwegyir-aggrey2023repairing} introduces {\em geometric repair}, adjusting a score function $s$ towards another score function $\hat{s}$ using a weighted average: $s_\lambda=(1-\lambda) \cdot s + \lambda \cdot \hat{s}$. Here, $\lambda$ determines the extent of adjustment. As \cite{kwegyir-aggrey2023repairing} explains, the term ``geometric'' is used because the adjustment process is a linear interpolation between the initial score s and the barycenter $\hat{s}$, which will result in $s_\lambda$. In our fair EM method, we apply geometric repair, adjusting $\lambda$ to minimize 
 \(\mathcal{U}(s_\lambda,\gamma)\), focusing on  \(\{\TPR, \FPR\}\) as \(\gamma\) values. AUC is primarily influenced by \TPR and \FPR, so concentrating on these values minimizes any adverse impact on AUC while ensuring fairness. Without a closed-form formula for \(\mathcal{U}(s_\lambda,\gamma)\), we employ a linear search across \(\lambda\) in [0,1].

\section{EXPERIMENTS}
\label{sec:experiment}

Our experiments focus on achieving two goals. Firstly, we investigate the effectiveness of traditional fairness metrics, including \DP, \EO, \EOD, and $\Delta \xAUC$, for measuring biases in EM. We find these measures often do not accurately reflect biases in EM, particularly when the matching thresholds vary. Instead, our metrics based on \DSP prove to be more effective. Secondly, we explore the impact of calibrating matching scores by aligning the score distributions of minority and majority groups towards their barycenters. This approach shows promising results in reducing \DSP without significantly affecting the accuracy of the corrected matching scores.


\starter{Experimental Setup:} Our approach builds upon existing state-of-the-art EM techniques, selecting four methods \hmatcher~\cite{fu2021hierarchical}, \ditto~\cite{li2021deep}, \dmatcher~\cite{mudgal2018deep}, and \hiergat~\cite{yao2022entity}. We use datasets from popular EM benchmarks: Amazon-Google (\amzgog), Walmart-Amazon (\walamz), DBLP-Google Scholar (\DBLPGogS), and \DBLPACM. We classify entities into minority and majority groups across various datasets, similar to previous research (e.g.,~\cite{nilforoushan2022entity,shahbazi2023through}). Specifically, in \amzgog, the ``manufacturer'' attribute's inclusion of ``Microsoft'' identifies a minority group. In \walamz, the minority group consists of entities categorized as ``printers'' within the ``category'' attribute. For \DBLPGogS, the ``venue'' attribute containing ``vldb j'' denotes the minority, while in \DBLPACM, entities are considered part of the minority if the ``authors'' attribute includes a female name. Comprehensive details on these datasets---including their statistics, the division between training and test sets, record and attribute counts, and the distribution of matched versus unmatched records---are available in our GitHub repository~\cite{mohammad}.\ignore{~\footnote{~\href{https://github.com/mhmoslemi2338/CaliFair-EM}{https://github.com/mhmoslemi2338/CaliFair-EM}}} The repository also contains additional experiments with other benchmarks and all the source code needed to reproduce the findings of our study.



\ignore{\ditto~ utilizes a pre-trained language model, enhanced through text summarization, data augmentation, and domain knowledge. \dmatcher~  offers a variety of methods, including RNN-based attention, latent semantic feature learning, and aggregation. \hmatcher~  extends \dmatcher~   with \mhm{dont know how to solve out of line issue }cross-attribute token alignment and attribute-aware attention, while \hiergat~employs a Hierarchical Graph Attention Transformer to address the ER challenge, learning contextual embeddings in the process. For pre-trained transformers and language models, we employed BERT (Bidirectional Encoder Representations from Transformers) ~\cite{devlin2018bert}. We utilized the codebases from \dmatcher~ ~\footnote{\href{https://github.com/anhaidgroup/deepmatcher}{https://github.com/anhaidgroup/deepmatcher}}, \texttt{EntityMatcher}~\footnote{\href{https://github.com/cascip/EntityMatcher}{https://github.com/cascip/EntityMatcher}}, \ditto~\footnote{\href{https://github.com/megagonlabs/ditto}{https://github.com/megagonlabs/ditto}}, and \hiergat~\footnote{\href{https://github.com/CGCL-codes/HierGAT}{https://github.com/CGCL-codes/HierGAT}} repositories, adapting their functionalities to suit our experimental requirements.}

\label{sec:biascurrent}

\starter{\DSP in Current Matching Methods:} Our first experiment assesses the matching methods' biases using \DSP. Table~\ref{tab:bias} shows the results in the ``before'' columns. The methods exhibit varying bias levels across the four datasets. \amzgog, \walamz, and \DBLPGogS demonstrate significant biases in all three \DSP measures (\DP, \EO, and \EOD), whereas \DBLPACM shows minimal biases. The AUC-based bias measure does not consistently identify these biases in methods with noticeable \DSP biases. For instance, neither \hiergat nor \hmatcher in \amzgog, nor \ditto, \hiergat, and \hmatcher in \DBLPGogS, are flagged by $\Delta \xAUC$. This inconsistency occurs because $\Delta \xAUC$ may overlook biases when \xAUC is nearly identical for different groups despite variations at specific thresholds. In general, \DP experiences more significant changes than \EOD and \EO. This is because \DP focuses on ensuring that both groups have identical distributions. By aligning both distributions to their barycenter, we achieve this objective. We also analyzed biases at two threshold values, $\tau=0.5$ and $0.9$, to support the theory that biases can vary significantly across thresholds. For instance, \EO and \EOD in \hmatcher with \amzgog are 0.3337 and 0.2995 at $\tau=0.5$, but drop to 0.1074 and 0.0946 at $\tau=0.9$, showing a significant decrease in biases at the higher threshold. However, biases do not always decrease at higher thresholds. For example, in \amzgog, \dmatcher's \EO and \EOD scores are 0.1720 and 0.1306 at $\tau=0.5$, increasing to 0.2084 and 0.1847 at $\tau=0.9$. This demonstrates that a comprehensive understanding of biases requires an examination across different thresholds, as enabled by \DSP.
 
{\em The experiment demonstrates that \DSP identifies biases in the current matching methods undetected by existing bias measures.}

\label{sec:calibration}
\starter{Score Calibration:} We investigated the impact of calibrating matching scores from existing methods to mitigate biases. We optimized parameter $\lambda$ for each bias measure, such as \DP and \EOD, recognizing that these measures can sometimes yield conflicting outcomes. The calibration results are presented in Table~\ref{tab:bias} in columns labeled ``after.'' These results demonstrate that calibration effectively reduces \DSP biases where there are significant biases, such as \amzgog, \walamz, \DBLPGogS, without increasing biases where the biases were insignificant, such as in \DBLPACM. This outcome results from the optimization that can only reduce biases as $\lambda=0$ gives the same initial biases. An important point from the results is that, although \DSP biases decrease $\Delta \xAUC$ does not necessarily follow suit, highlighting a distinct behavior between $\Delta \xAUC$ and \DSP.

\begin{figure}
         \centering\subfloat[\EO]{\label{fig:eo}
			{\includegraphics[width=0.48\linewidth]{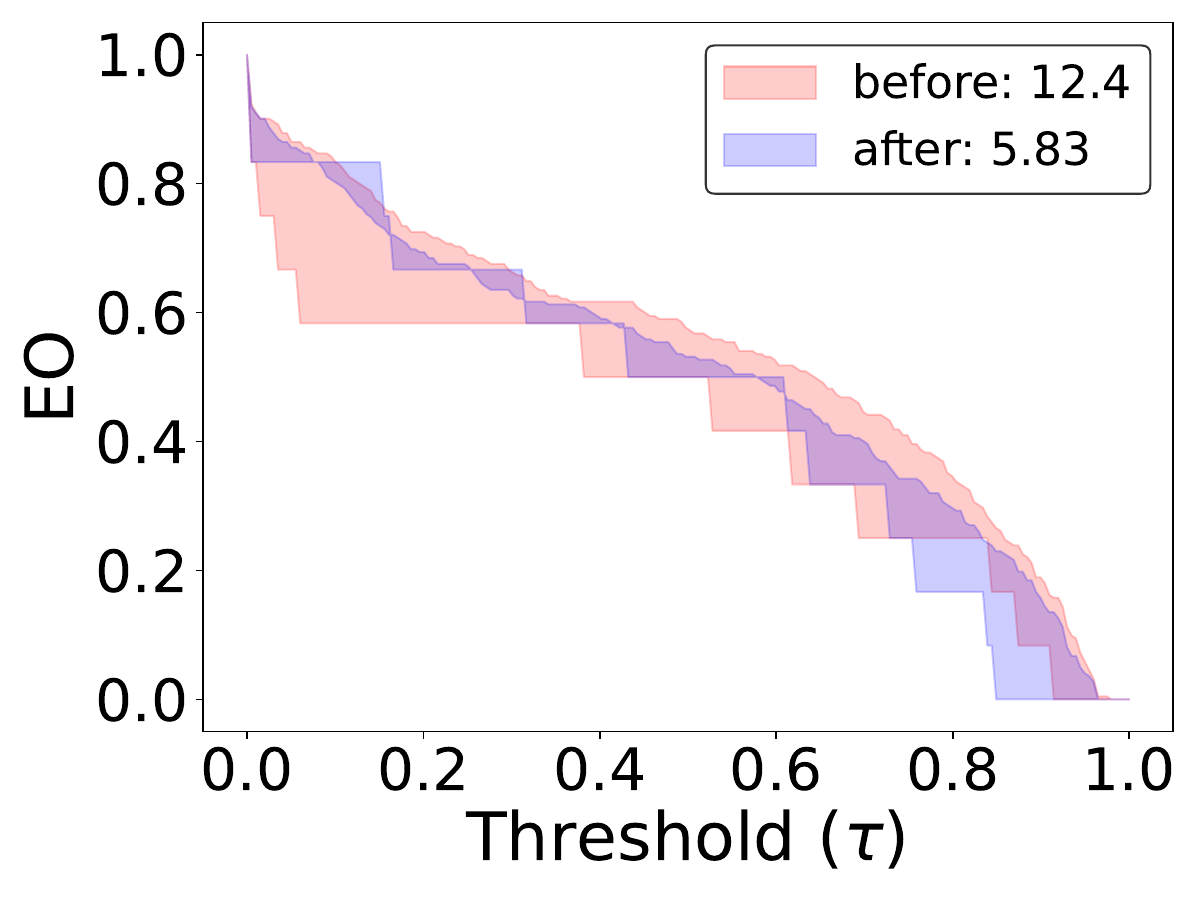}}}
		\hfill\subfloat[\EOD]{\label{fig:eod}
			{\includegraphics[width=0.48\linewidth]{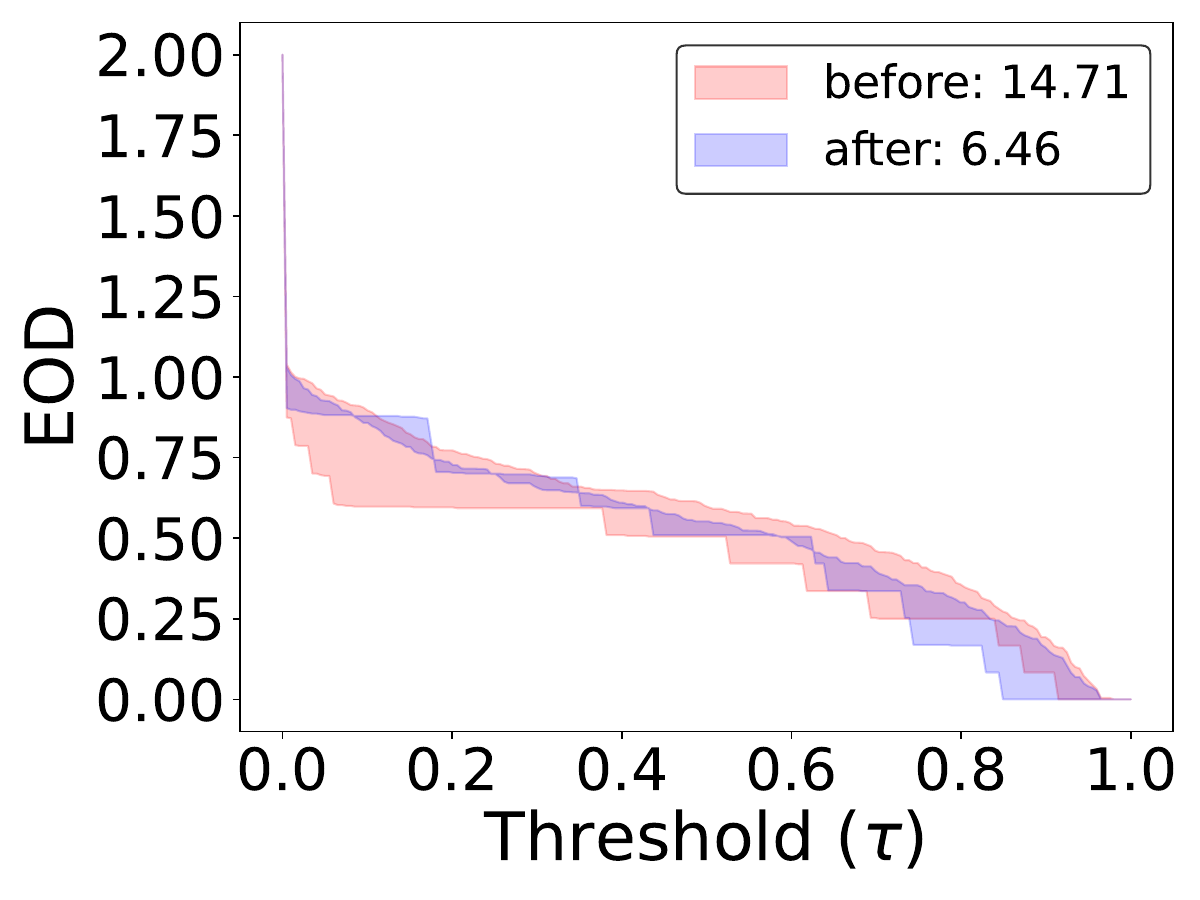}}}	\\
	         \centering\subfloat[\DP]{\label{fig:pr}
			{\includegraphics[width=0.48\linewidth]{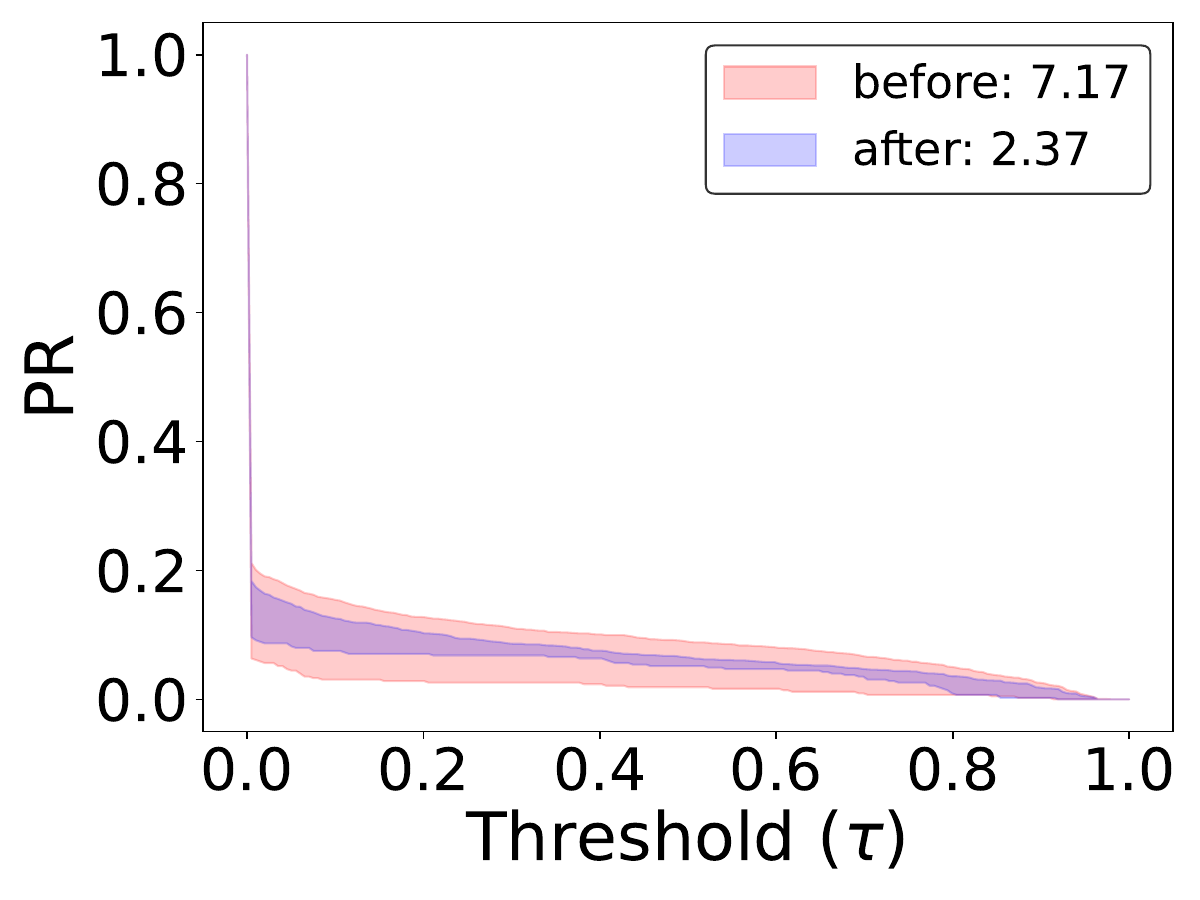}}}
		\hfill\subfloat[\AUC]{\label{fig:auc}
			{\includegraphics[width=0.48\linewidth]{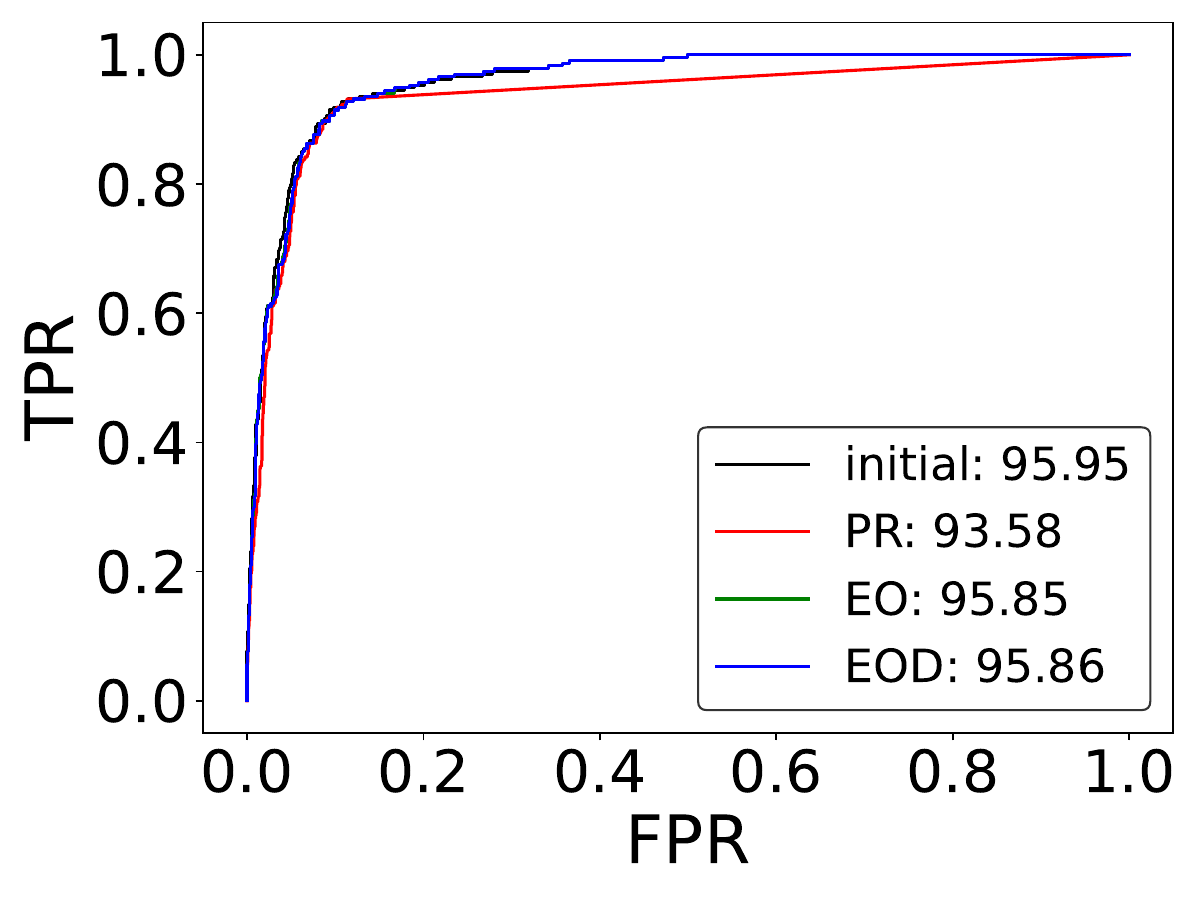}}}	
			\vspace{-3mm}
	\caption{DSP (Fig~\ref{fig:eo}-\ref{fig:pr}) and \AUC (Fig~\ref{fig:auc}), before and after score calibration, using \dmatcher and \walamz}\label{fig:allfigs}
 \vspace{-4mm}
\end{figure}

Figures~\ref{fig:eo} to \ref{fig:pr} illustrate the \DSP biases before and after the calibration. Areas highlighted in these figures denote biases at various thresholds (the difference between the curves for the minority and majority groups), with a reduction in these highlighted areas post-calibration indicating a bias decrease across thresholds. The figures also detail the \DSP biases, scaled by a factor of $\times 100$. Figure~\ref{fig:auc} compares the ROC of the matching score before and after calibration, labeled as ``initial'' and with tags \PR, \EOD, and \EO, respectively. The legend numbers represent AUC values. ``Initial'' is the initial AUC, while ``\PR'', ``\EOD'', and ``\EO'' are AUCs after calibration for each metric. The initial AUC and the AUCs for \EOD and \EO are nearly identical, resulting in overlapping curves that appear as a single blue curve. Despite a slight decrease in accuracy post-calibration, as indicated by the AUC values, the reduction is minimal. The more significant accuracy drop observed for \PR, which correlates with \DP, is because calibration modifies the scores of both groups to the barycenter. In contrast, \EO and \EOD adjustments are less drastic, merely moving scores toward the barycenter based on $\lambda$. Therefore, \DP calibration results in the most substantial deviation from the initial scores, leading to a larger impact on accuracy.


{\em The experiment indicates score calibration efficiently lowers \DSP in modern matching methods while maintaining high accuracy.}

\section{Conclusion}\label{sec:conclusion} 

In this paper, we have explored the evaluation of bias at various matching thresholds in EM methods, using \DSP to assess biases across these thresholds. Our experimental findings confirm the utility of \DSP in identifying biases that traditional bias metrics may miss. We also formulated the problem of fair EM, and we implemented a calibration technique as a post-processing debiasing approach that adjusts scores within groups towards a common barycenter to rescue biases in scores. This method has been shown to reduce biases while effectively maintaining high accuracy. Although our implementation yields promising results, future work is needed to address the challenge of achieving fair EM under strict fairness constraints, especially in providing guarantees regarding the accuracy of the calibrated scores to their original optimal values. A future approach will be reformulating the fair EM problem as an optimization problem with an objective that minimizes change on the score and also incorporates the fairness constraints in the objective function with the space of possible score functions beyond geometric repairs.

\bibliographystyle{ACM-Reference-Format}

\bibliography{ref}

\end{document}